\documentclass[runningheads]{llncs}
\usepackage{makecell}
\usepackage{multicol}
\usepackage{booktabs}
\usepackage{multirow}
\usepackage[T1]{fontenc}
\usepackage{tikz}
 \usepackage{amsmath} 
\usepackage{pgfplots}
\pgfplotsset{compat=newest}
\usepackage{pgfplotstable}
\usetikzlibrary{patterns}
\usepgfplotslibrary{polar}
\usepgfplotslibrary{groupplots}
\usepackage[table]{xcolor}
\usepackage{tcolorbox}
\usepackage{caption} 
\usepackage{url}

\usepackage[hidelinks]{hyperref}
\usepackage{xurl} 

\usepgfplotslibrary{fillbetween}
\usepackage{graphicx}
\definecolor{SciRed}{rgb}{0.84,0.37,0.00}       
\definecolor{SciBlue}{rgb}{0.00,0.45,0.70}      
\newcommand{\sigbar}[4]{%
\draw[thick]
  (axis cs:#1-0.18,#2) -- (axis cs:#1-0.18,#2+#3)
  -- (axis cs:#1+0.18,#2+#3)
  -- (axis cs:#1+0.18,#2);
\node[font=\scriptsize]
  at (axis cs:#1,#2+#3+0.035) {#4};
}

\newcommand{\blfootnote}[1]{%
  \begingroup
  \renewcommand{\thefootnote}{}%
  \footnote{#1}%
  \addtocounter{footnote}{-1}%
  \endgroup
}

\begin{document}
\title{From Many to Meaningful: Feature-Guided Zero-Shot Chronic Kidney Disease Screening using Large Language Models}
\titlerunning{Chronic Kidney Disease Screening using Large Language Models}
%
\author{Muhammad Ashad Kabir\inst{1}\orcidID{0000-0002-6798-6535} \and
Sirajam Munira\inst{2}} 
%
\authorrunning{M. A. Kabir et al.}
%
\institute{School of Computing, Mathematics and Engineering, Charles Sturt University, NSW, Australia 
\email{akabir@csu.edu.au}\\
\and
Department of Computer Science, Rensselaer Polytechnic Institute, NY, USA\\
\email{munirs@rpi.edu}}
\maketitle              

\blfootnote{%
\textbf{Publication notice.}
This is the author-prepared preprint of the following work:

\medskip
\noindent
Kabir, M.A., Munira, S.,
``From Many to Meaningful: Feature-Guided Zero-Shot Chronic Kidney Disease Screening Using Large Language Models,''
in \emph{Proceedings of the Artificial Intelligence in Medicine (AIME)},
Andreev, P., Van Woensel, W., Holmes, J., Sauré, A. (eds),
Lecture Notes in Computer Science, vol.~16748,
pp.~38--48.
Springer, Cham (2027).
\url{https://doi.org/10.1007/978-3-032-30710-1_5}.

\medskip
\noindent
Copyright notice for the published Version of Record:
\copyright~2027 The Author(s), under exclusive license to
Springer Nature Switzerland AG.

}

\begin{abstract}

Early screening of chronic kidney disease (CKD) is essential for preventing irreversible progression; however, many machine learning (ML)-based screening methods remain difficult to deploy in community and resource-limited screening settings due to their reliance on large labeled datasets, resource-intensive pathology tests, or high-dimensional clinical features, and limited robustness to population and distributional shifts. This study examines the feasibility of using large language models (LLMs) for early-stage CKD screening in a zero-shot setting, without dataset-specific training. We propose a feature-guided zero-shot framework that evaluates LLM performance using a selected set of clinically meaningful, readily available community-based features, rather than exhaustive clinical inputs.
Feature selection was guided by ML-based analysis to identify a compact, clinically relevant subset of variables. Tabular patient records were subsequently serialized into text using standardized prompt templates to enable zero-shot inference. The zero-shot performance of four LLMs (LLaMA-3, Qwen-3, Mistral, and GPT-4o-mini) was evaluated using both the full feature set and the selected subset. Generalizability was assessed across three heterogeneous CKD datasets spanning three countries. 
Across models and datasets, the selected feature set yielded consistent and statistically significant improvements in balanced accuracy and probability estimates, achieving performance levels suitable for screening purposes. These findings suggest that LLMs can support clinically meaningful, training-free CKD screening using minimal community-accessible patient features, offering a practical complement to conventional ML methods in real-world screening contexts.

\keywords{Chronic Kidney Disease  \and Large Language Model \and Zero-shot \and Tabular data \and Classification \and Feature guided}
\end{abstract}
\section{Introduction}
Chronic kidney disease (CKD) is a progressive condition characterized by sustained loss of kidney function or persistent kidney damage. If undetected, it can lead to end-stage renal disease requiring costly and lifelong treatment~\cite{levey2022chronic}. Globally, CKD affects approximately 850 million individuals, with a disproportionate burden in low- and middle-income countries, where access to early diagnosis and preventive care remains limited~\cite{stanifer2016chronic}. In South Asia, particularly in Bangladesh, community-based studies report high CKD prevalence and low awareness in rural and peri-urban populations~\cite{banik2021prevalence,sarker2021community}. Early-stage CKD (stages 1–3) is often asymptomatic, yet timely detection can significantly slow disease progression~\cite{levey2022chronic}.

Standard CKD screening relies on laboratory markers such as serum creatinine, estimated glomerular filtration rate (eGFR), and urine albumin-to-creatinine ratio, which require well-equipped facilities and trained personnel~\cite{levey2022chronic}. In many parts of South Asia, particularly rural and peri-urban areas, limited diagnostic infrastructure and financial constraints restrict access to these tests~\cite {stanifer2016chronic}. While conventional machine learning (ML) models have shown promising performance for CKD detection, they depend on large labeled datasets, predefined feature spaces, and retraining for new populations~\cite{sabanayagam2025artificial}. Such requirements reduce adaptability in heterogeneous, real-world screening contexts.

Recent advances in large language models (LLMs) have shown strong zero-shot generalization across diverse classification and reasoning tasks~\cite{brown2020language}. In nephrology, LLMs have been explored for various applications~\cite{hu2025large}, including patient education and personalized dietary advice~\cite{kairat2025benchmarking}, diagnostic support~\cite{syeda2024llm}, and medication management~\cite{sabanayagam2025artificial}. However, their use for systematic CKD screening -- particularly in zero-shot settings and based on non-diagnostic or readily accessible community-level features -- remains largely unexplored.

Motivated by this gap, we investigate whether LLMs can achieve clinically meaningful zero-shot CKD detection without any dataset-specific training or finetuning. This study makes three primary contributions: (i) we evaluate the zero-shot screening performance of four state-of-the-art LLMs (LLaMA-3, Qwen-3, Mistral, and GPT-4o-mini); (ii) we systematically compare the use of comprehensive feature sets with a compact subset of clinically meaningful variables, examining whether feature reduction improves or maintains screening performance; (iii) we assess generalizability by evaluating the approach on a community-based population and validating it on two additional datasets from different countries. Although the datasets do not share identical feature spaces, a harmonized subset of selected features was available across datasets. Across models and datasets, the selected feature subset achieved comparable or improved performance relative to using all available features. These findings suggest that meaningful feature selection can enhance zero-shot LLM-based CKD screening, which is critical for resource-constrained settings, and demonstrate cross-population robustness. 

\section{Methods}

\begin{figure}[!ht]
\includegraphics[width=\textwidth]{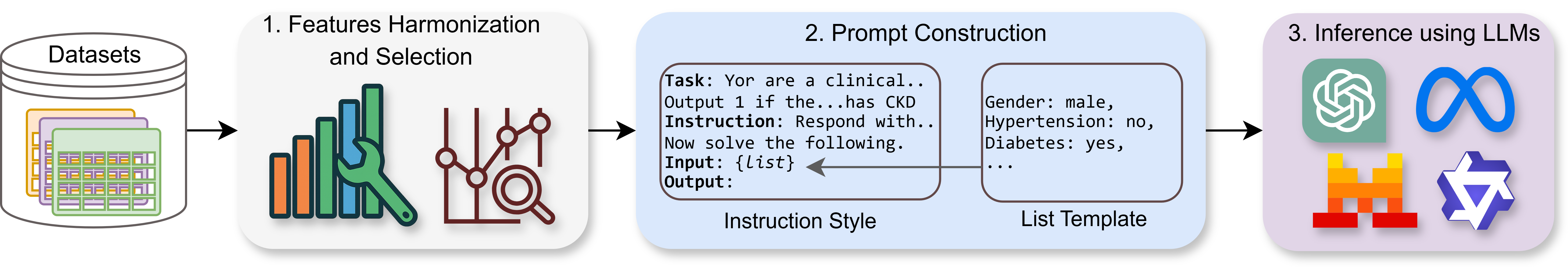}
\caption{Schematic overview of the zero-shot LLM pipeline for CKD detection} \label{fig:method}
\end{figure}

Fig.~\ref{fig:method} provides an overview of the proposed zero-shot LLM framework for CKD screening. The workflow consists of three main steps: (i) feature harmonization and selection, (ii) prompt construction, and (iii) inference using large language models. 

\subsection{Datasets}

We evaluated the proposed zero-shot LLM framework using three heterogeneous CKD datasets, summarized in Table~\ref{table:datasets}, representing both community-based (Dataset-1) and hospital-based populations (Dataset-2 and Dataset-3) across three countries: Bangladesh, India, and the UAE. To align with our research objective, gold-standard pathology tests used for direct CKD diagnosis, such as serum creatinine, urinary albumin, urinary creatinine, eGFR, and uACR, were excluded from all datasets. These markers are direct clinical identifiers of CKD. Therefore, including them would trivialize the prediction task. Moreover, such pathology tests are costly and not always accessible in low-resource or community settings. By excluding these features, we assess whether LLMs can detect CKD using more accessible and non-gold-standard clinical information. 

\begin{table}[!t]
    \centering
    \caption{Summary of CKD datasets used in this study}
    \label{table:datasets}
    {\footnotesize 
    \begin{tabular}{ll rrr c ccccc c}
        \hline
        \multirow{2}{*}{Dataset}  & \multirow{2}{*}{\makecell{Country}} & \multicolumn{3}{c}{Sample size} & \multirow{2}{*}{\makecell[c]{CKD\\[-4pt]Stage}} & \multicolumn{6}{c}{Considered Features}\\ 
        \cline{3-5}
        \cline{7-12}
        & & CKD & Non-CKD & Total & & SD & LH & MH & CE & Path* & Total\\
        \hline
        \hline
        Dataset-1 \cite{sarker2021community} & Bangladesh & 112 & 172 & 284 & 1--3 & 5 & 3 & 7 & 4 & 5 & 24 \\
        Dataset-2 \cite{chronic_kidney_disease_336} & India & 250 & 150 & 400 & -- & 1 & 1 & 3 & 5 & 13 & 23 \\
        Dataset-3 \cite{Al-Shamsi2018} & UAE & 56 & -- & 56 & 3--5 & 2 & 1 & 8 & 1 & 2 & 14\\
  
        \bottomrule
        \multicolumn{12}{l}{{\small\makecell[t l]{*Excluding gold-standard pathology tests for CKD diagnosis, such as serum\\creatinine, urinary albumin, urinary creatinine, eGFR, and uACR}}}
    \end{tabular}
    }
\end{table}

Dataset-1~\cite{sarker2021community} is a community-based CKD survey conducted in the Mirzapur, sub-district of Tangail, Bangladesh. This dataset was collected through age-stratified random sampling of adults aged 18 years and above under the Mirzapur demographic surveillance system. CKD was diagnosed based on reduced kidney function (eGFR) and/or elevated urine albumin-to-creatinine ratio (uACR) according to standard clinical guidelines. After removing incomplete records, the final dataset included 284 participants, consisting of 112 early-stage CKD cases (stages 1–3) and 172 non-CKD individuals. The dataset contains 24 variables covering socio-demographic (SD), lifestyle habits (LH), medical history (MH), clinical examination (CE), and pathology test (Path) features.

Dataset-2~\cite{chronic_kidney_disease_336} is the UCI 2015 CKD dataset, collected from a hospital in India and widely used as a benchmark for CKD detection research. It contains clinical records from 400 patients admitted over two months, including 250 CKD cases and 150 non-CKD controls. Following the removal of gold-standard pathology markers, 23 variables were retained across socio-demographic, lifestyle, medical history, clinical examination, and non-gold-standard pathology features. CKD stages were not reported in this dataset.

Dataset-3~\cite{Al-Shamsi2018} is the UAE CKD dataset derived from hospital records at Tawam Hospital in Al-Ain City, Abu Dhabi, UAE. It includes 56 patients diagnosed with moderate-to-advanced CKD (stages 3–5). CKD, with stages 1–2, was not clearly separated from healthy individuals in the original data; therefore, only confirmed CKD cases were used in our analysis. After excluding gold-standard tests and redundant features, 14 features were retained across demographic, lifestyle, medical history, clinical, and pathology.
\begin{table}[!t]
    \centering
        \caption{All features and selected features (underlined) retained after harmonization for each dataset.}
    \label{tab:datasetFeatures}
    \begin{tabular}{cp{10.5cm}}
    \hline
       Dataset  & Considered features\\
       \hline\hline
        Dataset-1 & \underline{Age}, \underline{Gender}, Illiterate, Occupation, Marital status, \underline{Sleeping duration}, Tobacco smoker, Smokeless tobacco, \underline{History of hypertension}, \underline{History of diabetes}, Heart disease, Stroke, Family history of diabetes, \underline{Family history of hypertension}, Family history of CKD, \underline{Body mass index}, Abdominal obesity, Undernutrition, \underline{Anemia}, \underline{Presence of red blood cells in urine}, Serum albumin, Hyper-cholesterolemia, HDL cholesterol, Hypertriglyceridemia\\
        \hline
        Dataset-2 & \underline{Age}, Blood pressure, Specific gravity, Albumin, Sugar, \underline{Presence of red blood cells in urine}, Pus cell, Pus cell clumps, Bacteria, Blood glucose random, Blood urea, Sodium, Potassium, Hemoglobin, Packed cell volume, White blood cell count, Red blood cell count, \underline{History of hypertension}, \underline{History of Diabetes}, Coronary artery disease, Appetite condition, Pedal edema, \underline{Anemia}\\\hline
        Dataset-3 & \underline{Age}, \underline{Gender}, History of heart disease, Vascular disease, History of dyslipidemia, \underline{History of hypertension}, \underline{History of diabetes}, Tobacco smoker, \underline{Body mass index}, Hyper-cholesterolemia, Consuming dyslipidemia medication, Consuming diabetes medication, Consuming hypertension medication, HbA1C\\
        \hline
    \end{tabular}
\end{table}

\subsection{Features Harmonization and Selection}

To ensure consistency across datasets, we performed systematic feature harmonization before model evaluation. Variables representing the same clinical concepts were harmonized across three datasets by standardizing both feature names and value encodings into a unified, semantically interpretable format to facilitate zero-shot LLM inference. This harmonization enabled consistent prompt construction and comparable evaluation across datasets. The complete list of harmonized features considered in this study is presented in Table~\ref{tab:datasetFeatures}.

Feature selection was informed by our prior machine learning analysis on the community-based Dataset-1~\cite{kabir2026community}, which includes diverse and accessible variables spanning socio-demographic characteristics, lifestyle factors, medical history, clinical examination findings, and selected pathology measures. In~\cite{kabir2026community}, ten complementary feature selection methods (including recursive feature elimination with multiple ML estimators) were applied to identify a compact subset of clinically meaningful and readily obtainable variables most predictive of early-stage CKD. The selected features are underlined in Table~\ref{tab:datasetFeatures}.
The selected features from Dataset-1 were subsequently mapped to Dataset-2 and Dataset-3 by identifying variables with equivalent clinical interpretation. As the datasets did not share identical feature spaces, only variables matching the selected features in Dataset-1 were retained as selected features in Datasets 2 and 3.

\subsection{Prompt Construction}
We represent each patient record (tabular instance) by serializing feature-value pairs into a structured list format (Appendix~\ref{sec:prompt}). This design follows prior findings demonstrating that list-based serialization preserves feature boundaries, reduces ambiguity, and performs comparably to free-text or sentence-style representations when applying LLMs to tabular inference~\cite{hegselmann2023tabllm}. The serialized representation enables the model to reason over individual attributes while preserving the original tabular semantics.

We adopt an instruction-style, role-conditioned prompting strategy (chat-style for GPT-4o-mini) in which each serialized record is embedded within a structured instruction–input–output format~\cite{wang2022selfinstruct}. The prompt (Appendix~\ref{sec:prompt}) explicitly specifies the task objective, assigns the model a functional role, defines the expected binary output space, and presents the serialized input record. This design is consistent with instruction-tuning paradigms used in modern LLM training, which have been shown to enhance zero-shot generalization~\cite{wei2021finetuned}. Unlike native chat-style templates, instruction-style prompts avoid reliance on model-specific control tokens, improving robustness and cross-model generalizability. Furthermore, explicit output constraints reduce generation variability and support more deterministic predictions, which is advantageous for zero-shot binary classification of structured clinical data.

The prompt text was formulated by adapting instruction templates aligned with model instruction-tuning paradigms and iteratively refined through empirical validation, consistent with prior work on instruction-based prompt optimization in in-context learning settings~\cite{cui2025see}.

\subsection{Inference using LLMs}
We evaluated four large language models, LLaMA-3 8B, Qwen-3 8B, Mistral 7B, and GPT-4o-mini, selected to represent architectural diversity, variation in parameter scale, and differences in training paradigms (open-weight vs. proprietary). These models have been pretrained on large-scale, heterogeneous corpora that include biomedical and clinical knowledge, making them suitable for zero-shot reasoning in healthcare-related tasks such as CKD screening. Additionally, all selected models expose token-level log-probabilities, enabling probabilistic classification based on likelihood comparisons over predefined label tokens—an important capability not consistently available in many LLMs.

Inference was conducted in a zero-shot setting without task-specific training or parameter updates. For each patient record, the constructed instruction-style prompt was provided to the model, and binary predictions (CKD vs. non-CKD) were derived using log-probability scores over the constrained output space. Deterministic decoding was applied where supported to ensure reproducibility. Each record was processed independently, and no in-context examples were provided.

\section{Results}

\begin{figure}[!t]
\centering
\begin{tikzpicture}
\begin{groupplot}[
    ybar,
    enlargelimits=0.15,
    footnotesize,
    ymin=0.14, ymax=1.2,
    ytick={0,0.2,0.4,0.6,0.8, 1},
    xtick={0,1,2,3},
    xticklabel style={rotate=45, anchor=east, align=center},
    legend columns=-1,
    nodes near coords ,
    every node near coord/.append style={rotate=90,anchor=west, /pgf/number format/precision=2,  font=\scriptsize}, 
    name=mygroup,
            xticklabels={
        LLaMA-3,
        Qwen-3,
        Mistral,
        GPT-4o
    },
    group style={
        group size=3 by 1,
        xlabels at=edge bottom,
        horizontal sep=.9cm,   
        ylabels at=edge left
    }
]

\nextgroupplot[
    width=0.37\textwidth,
    height=5cm,
    bar width=6pt,
    ylabel={Balanced Accuracy},
    xlabel={(a) Dataset-1},
    legend style = { text=black, column sep = 1pt, legend columns = -1, legend to name = grouplegend}
]

\addplot[pattern=dots, pattern color=SciRed, draw=SciRed, error bars/.cd, y dir=both, y explicit] 
    coordinates { (0, 0.5)
    (1, 0.6206) (2, 0.5552) (3, 0.6972)};
\addlegendentry{All features (All)}
\addplot[pattern=north east lines, pattern color=SciBlue, draw=SciBlue,  error bars/.cd, 
          y dir=both, y explicit] 
    coordinates {      
        (0, 0.5419) 
        (1, 0.7829) 
        (2, 0.7721) 
        (3, 0.7708)
        };
\addlegendentry{Selected features (S)}
\sigbar{0}{1.23}{0.04}{***} 
\sigbar{1}{1.23}{0.04}{***} 
\sigbar{2}{1.23}{0.04}{***} 
\sigbar{3}{1.23}{0.04}{***} 
\nextgroupplot[
     width=0.38\textwidth,
    height=5cm,
    bar width=6pt,
    xlabel={(b)} Dataset-2,
    ]
\addplot[pattern=dots, pattern color=SciRed, draw=SciRed,  error bars/.cd,
          y dir=both, y explicit] coordinates {(0,0.50195) (1,0.81605) (2,0.53469) (3,0.85567)};
\addplot[pattern=north east lines, pattern color=SciBlue, draw=SciBlue,  error bars/.cd, 
          y dir=both, y explicit] coordinates {(0,0.76455) (1,0.81237) (2,0.81215) (3,0.88001)};

\sigbar{0}{1.23}{0.04}{***} 
\sigbar{1}{1.23}{0.04}{**}
\sigbar{2}{1.23}{0.04}{***}
\sigbar{3}{1.23}{0.04}{\textit{n.s.}}

\nextgroupplot[
    width=0.38\textwidth,
    height=5cm,
    bar width=6pt,
    xlabel={(c)} Dataset-3,
    ]
\addplot[pattern=dots, pattern color=SciRed, draw=SciRed,  error bars/.cd,
          y dir=both, y explicit] coordinates {(0,0.0) (1,0.9288) (2,1.0) (3,0.96477)};
\addplot[pattern=north east lines, pattern color=SciBlue, draw=SciBlue,  error bars/.cd, 
          y dir=both, y explicit] coordinates {(0,0.0) (1,0.97477) (2,0.96477) (3,0.94648)};
\sigbar{0}{1.23}{0.04}{***}
\sigbar{1}{1.23}{0.04}{***}
\sigbar{2}{1.23}{0.04}{***}
\sigbar{3}{1.23}{0.04}{**}
\end{groupplot}
\node at ($(current bounding box.north) + (0cm, 0.6cm)$) {\ref*{grouplegend}};
\end{tikzpicture}
\caption{Comparison of \textit{all} versus \textit{selected} features and the performance of different LLMs. Bars report balanced accuracy, while statistical significance between all- and selected-feature settings is assessed using a paired permutation test on per-sample Brier loss, with pairing over shared test samples. Significance levels are indicated by asterisks: *$p<0.05$, **$p<0.01$ and ***$p<0.001$.} 
\label{fig:allvsSelected}
\end{figure}
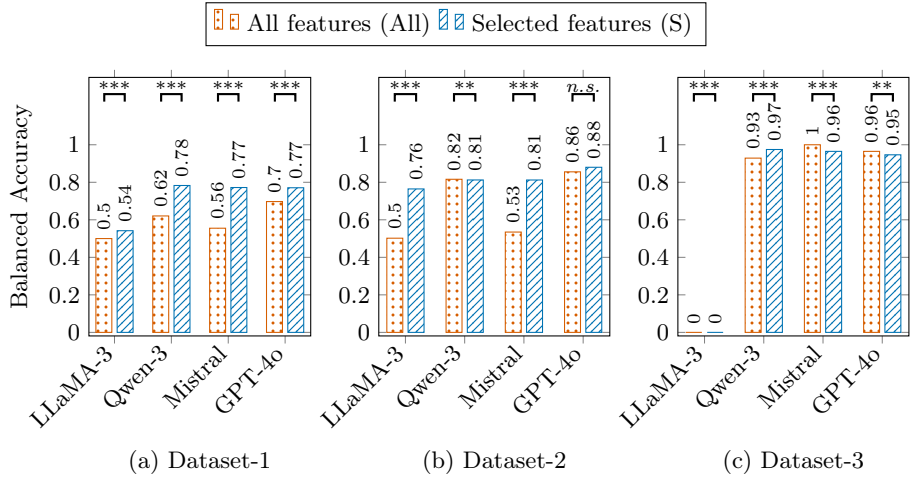

Fig.~\ref{fig:allvsSelected} compares the use of all available features versus the selected feature subset across LLMs and datasets using balanced accuracy, with statistical significance assessed using a paired permutation test on per-sample Brier loss to provide a distribution-free, paired comparison of probabilistic predictions. Table~\ref{tab:allResult} reports the detailed zero-shot performance metrics, including balanced accuracy, macro AUROC, macro F1-score, and Brier score, along with 95\% confidence intervals estimated using nonparametric bootstrap resampling at the sample level using percentile intervals to quantify metric uncertainty.

\begin{table}[!ht]
\caption{Zero-shot performance of LLMs using all features and selected features, reported as balanced accuracy, macro AUROC, macro F1-score, and Brier loss with 95\% confidence intervals computed using nonparametric bootstrap.}
\label{tab:allResult}
\begin{tabular}{cll cccc}
\hline
Dataset & Model & FS& B. Accuracy (\%) & Macro AUROC & Macro F1 (\%) & Brier loss ($\downarrow$) \\
\hline\hline
\multirow{8}{*}{\rotatebox{90}{Dataset-1}}
& Qwen-3 & All & $62.04_{(57.14-66.79)}$ & $0.735_{(0.676\text{--}0.794)}$ & $55.90_{(50.17-61.67)}$ & $0.253_{(0.227-0.280)}$ \\
& & S & $78.32_{(73.43-83.42)}$ & $0.836_{(0.786-0.884)}$ & $78.63_{(73.71-83.62)}$ & $0.161_{(0.131-0.192)}$ \\ \cmidrule{2-7}

& LLaMa-3 & All & $50.00_{(50.00-50.00)}$ & $0.749_{(0.689-0.804)}$ & $37.69_{(35.31-39.83)}$ & $0.264_{(0.232-0.297)}$ \\
& & S & $54.26_{(51.30-57.31)}$ & $0.816_{(0.766-0.865)}$ & $47.58_{(42.32-53.17)}$ & $0.207_{(0.181-0.235)}$ \\ \cmidrule{2-7}

& GPT-4o & All & $69.69_{(65.19-73.94)}$ & $0.803_{(0.751-0.855)}$ & $64.21_{(58.60-69.39)}$ & $0.330_{(0.280-0.382)}$ \\
& & S & $77.11_{(72.48-81.79)}$ & $0.851_{(0.808-0.892)}$ & $75.77_{(70.93-80.51)}$ & $0.212_{(0.170-0.256)}$ \\ \cmidrule{2-7}

& Mistral & All & $55.53_{(53.23-58.08)}$ & $0.794_{(0.739-0.848)}$ & $39.61_{(34.33-45.18)}$ & $0.505_{(0.448-0.558)}$ \\
&  & S & $77.27_{(72.38-82.16)}$ & $0.838_{(0.791-0.883)}$ & $76.51_{(71.66-81.34)}$ & $0.212_{(0.169-0.255)}$ \\
\hline
\multirow{8}{*}{\rotatebox{90}{Dataset-2}}
 & GPT-4o & All & $85.57_{(81.85-89.25)}$ & $0.925_{(0.896-0.950)}$ & $85.40_{(81.73-89.01)}$ & $0.131_{(0.098-0.165)}$ \\ 
& & S & $88.00_{(85.34-90.54)}$ & $0.943_{(0.921-0.964)}$ & $84.82_{(81.30-88.12)}$ & $0.145_{(0.114-0.178)}$ \\ \cmidrule{2-7}
& Qwen-3 & All & $81.61_{(78.60-84.49)}$ & $0.901_{(0.866-0.934)}$ & $76.95_{(72.69-80.98)}$ & $0.170_{(0.150-0.191)}$ \\
& & S & $81.24_{(78.14-84.29)}$ & $0.843_{(0.796-0.885)}$ & $76.50_{(72.43-80.67)}$ & $0.207_{(0.171-0.244)}$ \\ \cmidrule{2-7}
& Mistral & All & $53.47_{(51.13-56.05)}$ & $0.822_{(0.779-0.860)}$ & $46.92_{(42.48-51.45)}$ & $0.341_{(0.297-0.388)}$ \\
& & S & $85.22_{(82.35-88.04)}$ & $0.932_{(0.908-0.956)}$ & $81.40_{(77.46-85.16)}$ & $0.179_{(0.143-0.216)}$ \\ \cmidrule{2-7}
& LLaMa-3 & All & $50.20_{(50.00-50.62)}$ & $0.923_{(0.895-0.948)}$ & $27.68_{(25.09-30.32)}$ & $0.326_{(0.304-0.348)}$ \\
& & S & $76.45_{(73.22-79.67)}$ & $0.849_{(0.810-0.887)}$ & $70.46_{(65.92-74.99)}$ & $0.212_{(0.186-0.239)}$ \\

\hline
\multirow{8}{*}{\rotatebox{90}{Dataset-3}}
& Mistral & All & $100.0_{(100.0-100.0)}$ & -- & $100.0_{(100.0-100.0)}$ & $0.003_{(0.000-0.010)}$ \\
&  & S & $96.48_{(91.07-100.0)}$ & -- & $56.40_{(47.66-100.0)}$ & $0.035_{(0.000-0.088)}$ \\
\cmidrule{2-7}
& GPT-4o & All & $96.48_{(91.07-100.0)}$ &--& $56.40_{(47.66-100.0)}$ & $0.035_{(0.000-0.088)}$ \\
& & S & $94.65_{(87.50-100.0)}$ & -- & $50.84_{(46.67-100.0)}$ & $0.054_{(0.002-0.122)}$ \\
\cmidrule{2-7}
& Qwen-3 & All & $92.89_{(85.71-98.21)}$ & -- & $49.19_{(46.15-49.55)}$ & $0.084_{(0.031-0.152)}$ \\
&  & S & $96.48_{(91.07-100.0)}$ & -- & $56.40_{(47.66-100.0)}$ & $0.036_{(0.001-0.089)}$ \\
\cmidrule{2-7}
& LLaMa-3 & All & $0.00_{(0.00-0.00)}$ & -- & $0.00_{(0.00-0.00)}$ & $0.670_{(0.663-0.678)}$ \\
& & S & $0.00_{(0.00-0.00)}$ & -- & $0.00_{(0.00-0.00)}$ & $0.489_{(0.468-0.514)}$ \\

\hline
\multicolumn{7}{l}{B. Accuracy: Balanced accuracy, FS: Feature set, All: All features, S: Selected features}\\
\end{tabular}
\end{table}

For Dataset-1, the selected feature set consistently yields higher balanced accuracy across all models, with statistically significant improvements ($p<0.001$). The largest gains are observed for Qwen-3 and Mistral, where balanced accuracy increases from 0.62 to 0.78 and from 0.56 to 0.77, respectively. GPT-4o-mini improves from 0.70 to 0.77, while LLaMA-3 shows a modest but significant increase from 0.50 to 0.54. These improvements are accompanied by higher macro AUROC and macro F1-scores and lower Brier loss, indicating enhanced discrimination and calibration. For example, Qwen-3's AUROC increases from 0.735 to 0.836, macro F1-score from 55.90\% to 78.63\%, and Brier loss decreases from 0.253 to 0.161. The bootstrap-derived 95\% confidence intervals remain well separated in settings with large performance gains, reinforcing the statistical stability of the observed improvements. Similar patterns are observed across the other models, suggesting that restricting inputs to clinically selected features improves both predictive performance and probabilistic reliability.

For Dataset-2, selected features again yield improved or comparable balanced accuracy across all models, with statistically significant gains observed in several cases. The most substantial improvement is seen for Mistral, where balanced accuracy increases from 0.53 to 0.81 ($p<0.001$). LLaMA-3 also demonstrates a marked improvement from 0.50 to 0.76. Qwen-3 maintains strong performance under both feature configurations, indicating robustness to feature redundancy. While GPT-4o-mini shows a slight increase, that difference is not statistically significant. These performance patterns are consistent with changes in AUROC, F1-score, and Brier loss. For example, Mistral's AUROC increases from 0.822 to 0.932, and its Brier loss decreases from 0.341 to 0.179, indicating improved discrimination and probabilistic calibration when using selected features.

For Dataset-3, a different pattern is observed, likely reflecting its focus on moderate-to-advanced CKD cases. LLaMA-3 yields a balanced accuracy of 0 under both feature configurations, indicating an inability to identify CKD cases, although the difference between all and selected features remains statistically significant in terms of Brier loss. The most notable improvement is observed for Qwen-3, where balanced accuracy increases from 0.93 to 0.97 with statistical significance ($p<0.001$). In contrast, Mistral and GPT-4o-mini show slight decreases when using selected features, from 1.00 to 0.96 and from 0.96 to 0.95, respectively; however, overall performance remains high. These patterns are also reflected in AUROC and Brier scores. For example, Qwen-3's Brier loss decreases from 0.084 to 0.036, indicating improved probabilistic calibration in selected features, while Mistral and GPT-4o-mini maintain strong discrimination despite marginal reductions in balanced accuracy.

Overall, the results consistently indicate that the selected feature set improves or maintains balanced accuracy, AUROC, F1-score, and probabilistic calibration across datasets, despite differences in feature space composition. This consistency across heterogeneous cohorts supports the robustness of the approach and suggests that zero-shot LLM-based CKD screening can generalize across populations using a compact, clinically harmonized set of features.

\section{Discussions}
The most pronounced gains were observed in early-stage community screening (Dataset-1), where reducing redundant or less informative inputs substantially improved predictive stability and calibration. This finding suggests that in settings where disease signals are subtle, compact, and clinically meaningful features may enhance zero-shot reasoning by reducing noise and guiding the model toward salient risk factors. In hospital-based cohorts (Dataset-2), feature selection yielded particularly great improvements for smaller or open-weight models, whereas higher-capacity models such as GPT-4o-mini demonstrated relative robustness to feature redundancy. In the advanced-stage CKD cohort (Dataset-3), performance remained high under both feature configurations, indicating that feature reduction does not compromise discrimination when disease signals are more pronounced.

Our previous supervised ML study using the same selected feature subset reported a best balanced accuracy of 90.4\% on Dataset-1 using 10-fold cross-validation and 79.33\% when trained on Dataset-1 and externally tested on Dataset-2~\cite{kabir2026community}. In comparison, the zero-shot LLM approach achieved up to 78.32\% on Dataset-1 (Qwen-3) and 88\% on Dataset-2 (GPT-4o-mini) without any dataset-specific training, demonstrating competitive performance with strong practical applicability.

Collectively, these results suggest that feature-guided prompting enhances the reliability and stability of zero-shot LLM-based screening, particularly in early-stage and resource-constrained settings. While larger models appear less sensitive to redundant inputs, a compact and clinically grounded feature set provides clearer and more consistent performance gains across diverse model architectures without sacrificing predictive quality.

Several limitations should be acknowledged. First, the datasets are moderate in size and limited to three geographic regions, which may restrict generalizability. Second, zero-shot prompting relies on prompt design and instruction tuning characteristics, which may vary across future model releases. Third, Dataset-3 contains only CKD-positive cases, limiting full binary evaluation in that context.

\section{Conclusions and Future Work}
This study demonstrates that LLMs can support clinically meaningful zero-shot CKD screening using a compact, harmonized feature set without dataset-specific training or retraining. Across heterogeneous community- and hospital-based cohorts, restricting inputs to clinically selected features consistently improved or preserved accuracy, discrimination, and probabilistic calibration.

Future work should evaluate larger and more diverse cohorts, explore structured uncertainty estimation for screening thresholds, and investigate few-shot or retrieval-augmented extensions to further enhance robustness. Integration into community screening workflows and prospective validation in real-world settings will be critical to assess clinical impact.

\begin{credits}
\subsubsection{\ackname} Authors thank Dr. Mohammad Habibur Rehman Sarker (ICDDRB) for sharing Dataset-1.
\end{credits}
\appendix
\renewcommand{\thetable}{A\arabic{table}}
\setcounter{table}{0}
\section{Zero-Shot Prompt Template}\label{sec:prompt}
\captionof{table}{Example of the instruction-style prompt used for zero-shot inference.}
\label{tab:prompt_templates_xlp_baseline}
\begin{tcolorbox}
{\scriptsize
Task: You are a clinical classification model. Classify whether a patient has chronic kidney disease (CKD) based on the patient input features. 1 indicates CKD and 0 indicates non-CKD.   

Output 1 if the patient has CKD.
Output 0 if the patient does not have CKD.

Instruction: Respond with exactly one token: 0 or 1.

Now solve the following.

Input:

Patient features:

History of diabetes: Yes,

History of hypertension: No,

...

Output:
}
\end{tcolorbox}
%
%
\bibliographystyle{splncs04}
\bibliography{reference}

@inproceedings{syeda2024llm,
  title={LLM-based kidney disease diagnostic framework for Pathologists},
  author={Syeda, Masooma Zehra and Bukhari, Syed Usama Khalid and Hussain, Maqbool and Khan, Wajahat Ali and Shah, Syed Sajid Hussain},
  booktitle={46th Annual International Conference of the IEEE Engineering in Medicine and Biology Society},
  year={2024},
  organization={IEEE}
}

@article{hu2025large,
  title={Large language models in nephrology: applications and challenges in chronic kidney disease management},
  author={Hu, Yongzheng and Liu, Jianping and Jiang, Wei},
  journal={Renal Failure},
  volume={47},
  number={1},
  year={2025},
  publisher={Taylor \& Francis}
}

@article{sarker2021community,
  title={Community-based screening to determine the prevalence, health and nutritional status of patients with CKD in rural and peri-urban Bangladesh},
  author={Sarker, Mohammad Habibur Rahman and Moriyama, Michiko and Rashid, Harun Ur and Chisti, Mohammod Jobayer and Rahman, Md Moshiur and Das, Sumon Kumar and Uddin, Aftab and Saha, Samir Kumar and Arifeen, Shams El and Ahmed, Tahmeed and  Faruque, Asg},
  journal={Therapeutic advances in chronic disease},
  volume={12},
  year={2021},
  publisher={SAGE}
}

@misc{chronic_kidney_disease_336,
  author       = {Rubini, L. and Soundarapandian, P. and Eswaran, P.},
  title        = {{Chronic Kidney Disease}},
  year         = {2015},
  howpublished = {UCI Machine Learning Repository},
  note         = {\url{https://doi.org/10.24432/C5G020}}
}

@inproceedings{wei2021finetuned,
title={Finetuned Language Models are Zero-Shot Learners},
author={Jason Wei and Maarten Bosma and Vincent Zhao and Kelvin Guu and Adams Wei Yu and Brian Lester and Nan Du and Andrew M. Dai and Quoc V Le},
booktitle={International Conference on Learning Representations},
year={2022},
}

@inproceedings{wang2022selfinstruct,
  title={Self-instruct: Aligning language models with self-generated instructions},
  author={Wang, Yizhong and Kordi, Yeganeh and Mishra, Swaroop and Liu, Alisa and Smith, Noah A and Khashabi, Daniel and Hajishirzi, Hannaneh},
  booktitle={Proceedings of the 61st annual meeting of the Association for Computational Linguistics},
  pages={13484--13508},
  year={2023}
}

@inproceedings{hegselmann2023tabllm,
  title={{TabLLM}: Few-shot classification of tabular data with large language models},
  author={Hegselmann, Stefan and Buendia, Alejandro and Lang, Hunter and Agrawal, Monica and Jiang, Xiaoyi and Sontag, David},
  booktitle={International conference on artificial intelligence and statistics},
  pages={5549--5581},
  year={2023}
}

@article{kabir2026community,
title = {Community-based early-stage chronic kidney disease screening using explainable machine learning for low-resource settings},
journal = {International Journal of Medical Informatics},
volume = {214},
year = {2026},
author = {Muhammad Ashad Kabir and Sirajam Munira and Dewan Tasnia Azad and Saleh Mohammed Ikram and Mohammad Habibur Rahman Sarker and Syed Manzoor Ahmed Hanifi},
doi = {10.1016/j.ijmedinf.2026.106435},
}

@article{banik2021prevalence,
  title={Prevalence of chronic kidney disease in Bangladesh: a systematic review and meta-analysis},
  author={Banik, Sujan and Ghosh, Antara},
  journal={International Urology and Nephrology},
  volume={53},
  number={4},
  pages={713--718},
  year={2021},
  publisher={Springer}
}

@article{levey2022chronic,
  title={Chronic kidney disease},
  author={Levey, Andrew S and Coresh, Josef},
  journal={The Lancet},
  volume={399},
  number={10391},
  year={2022},
  publisher={Elsevier},
}

@inproceedings{cui2025see,
  title={{SEE}: Strategic exploration and exploitation for cohesive in-context prompt optimization},
  author={Cui, Wendi and Zhang, Jiaxin and Li, Zhuohang and Sun, Hao and Lopez, Damien and Das, Kamalika and Malin, Bradley A and Kumar, Sricharan},
  booktitle={Proceedings of the 63rd Annual Meeting of the Association for Computational Linguistics},
  pages={29575--29627},
  year={2025}
}

@article{kairat2025benchmarking,
  title={Benchmarking {ChatGPT} and Other Large Language Models for Personalized Stage-Specific Dietary Recommendations in Chronic Kidney Disease},
  author={Kairat, Makpal and Adilmetova, Gulnoza and Ibraimova, Ilvira and Gaipov, Abduzhappar and Varol, Huseyin Atakan and Chan, Mei-Yen},
  journal={Journal of Clinical Medicine},
  volume={14},
  number={22},
  year={2025},
  publisher={MDPI}
}

@article{sabanayagam2025artificial,
  title={Artificial intelligence in chronic kidney disease management: a scoping review},
  author={Sabanayagam, Charumathi and Banu, Riswana and Lim, Cynthia and Tham, Yih Chung and Cheng, Ching-Yu and Tan, Gavin and Ekinci, Elif and Sheng, Bin and McKay, Gareth and Shaw, Jonathan E and others},
  journal={Theranostics},
  volume={15},
  number={10},
  pages={4566--4578},
  year={2025}
}

@article{stanifer2016chronic,
  title={Chronic kidney disease in low-and middle-income countries},
  author={Stanifer, John W and Muiru, Anthony and Jafar, Tazeen H and Patel, Uptal D},
  journal={Nephrol. Dial. Transplant.},
  volume={31},
  number={6},
  pages={868--874},
  year={2016},
  publisher={Oxford University Press}
}

@inproceedings{brown2020language,
author = {Brown, Tom B. and Mann, Benjamin and Ryder, Nick and Subbiah, Melanie and Kaplan, Jared and Dhariwal, Prafulla and Neelakantan, Arvind and Shyam, Pranav and Sastry, Girish and Askell, Amanda and others},
title = {Language models are few-shot learners},
 volume = {33},
year = {2020},
booktitle = {Advances in Neural Information Processing Systems},
articleno = {159},
numpages = {25},
location = {Vancouver, BC, Canada},
}

@article{Al-Shamsi2018,
	title={Chronic kidney disease in patients at high risk of cardiovascular disease in the United Arab Emirates: A population-based study},
    journal = {PLOS ONE},
    publisher = {Public Library of Science},
	author={S Al-Shamsi and D Regmi and R D Govender},
number = {6},
  volume = {13},
	year={2018}
}
%




\end{document}